\documentclass[conference]{IEEEtran}
\IEEEoverridecommandlockouts
\usepackage{cite}
\usepackage{amsmath,amssymb,amsfonts}
\usepackage{algorithmic}
\usepackage{graphicx}
\usepackage{textcomp}
\usepackage{xcolor}
\usepackage{float}
\usepackage{hyperref}
\usepackage{algorithm}
\usepackage{amssymb, amsfonts}
\usepackage{makecell}
\usepackage{tikz}
\usetikzlibrary{positioning, shapes, arrows.meta, fit, backgrounds}
\usepackage{cite}
\usepackage{graphicx}
\usepackage{amsmath}
\def\BibTeX{{\rm B\kern-.05em{\sc i\kern-.025em b}\kern-.08em
    T\kern-.1667em\lower.7ex\hbox{E}\kern-.125emX}}
\begin{document}

\title{Learning Traffic Anomalies from \\ Generative Models on Real-Time Observations}

\author{\IEEEauthorblockN{Fotis I. Giasemis* \thanks{* Corresponding author}}
\IEEEauthorblockA{\textit{LIP6, LPNHE} \\
\textit{Sorbonne Université}\\
\textit{CNRS, IN2P3}\\
Paris, France\\
Fotis.Giasemis@cern.ch}
\and
\IEEEauthorblockN{Alexandros Sopasakis}
\IEEEauthorblockA{\textit{Department of Mathematics} \\
\textit{Lund University}\\
Lund, Scania, Sweden \\
Alexandros.Sopasakis@math.lth.se}
}

\maketitle
\thispagestyle{empty}

\begin{abstract}
Accurate detection of traffic anomalies is crucial for effective urban traffic management and congestion mitigation. We use the Spatiotemporal Generative Adversarial Network (STGAN) framework combining Graph Neural Networks and Long Short-Term Memory networks to capture complex spatial and temporal dependencies in traffic data. We apply STGAN to real-time, minute-by-minute observations from 42 traffic cameras across Gothenburg, Sweden, collected over several months in 2020. The images are processed to compute a flow metric representing vehicle density, which serves as input for the model. Training is conducted on data from April to November 2020, and validation is performed on a separate dataset from November 14 to 23, 2020. Our results demonstrate that the model effectively detects traffic anomalies with high precision and low false positive rates. The detected anomalies include camera signal interruptions, visual artifacts, and extreme weather conditions affecting traffic flow.
\end{abstract}


\IEEEpeerreviewmaketitle

\section{Introduction}

Urban traffic management systems rely heavily on the effective detection of traffic anomalies to prevent congestion and reduce accidents \cite{Zhao2019}. Rapid urbanization and the increasing complexity of traffic networks have rendered traditional statistical methods \cite{Yin2022, sopasakis2016information, Benarmas2023} insufficient for accurate traffic forecasting.
The high-dimensional and non-linear nature of traffic data necessitates advanced modeling techniques capable of capturing both spatial and temporal dependencies \cite{Patil2022, alperovich2008stochastic,  Sakurada2015}.

Recent advancements in artificial intelligence, particularly deep learning, have shown promise in improving traffic forecasting accuracy by leveraging large-scale data and computational resources \cite{Zhang2022}. These methods include graph-based deep learning models that efficiently handle spatial-temporal correlations and diffusion models that transform traffic forecasting into a conditional image generation task \cite{Lv2015}. Furthermore, innovative approaches that incorporate feature engineering without relying on real-time data have been developed to address challenges in urban traffic flow prediction \cite{Deng2022}.

Graph Neural Networks (GNNs) have emerged as powerful tools for modeling graph-structured data, effectively capturing spatial dependencies in traffic networks \cite{Wu2020, Zhang2022}. Meanwhile, Recurrent Neural Networks (RNNs), such as Long Short-Term Memory (LSTM) networks, are renowned for their ability to model temporal patterns \cite{Hochreiter1997, Sopasakis2019}. However, standalone LSTMs often fall short in exploiting the inherent spatial dependencies present in traffic data \cite{Yu2018}.

In this research, we use the Spatiotemporal Generative Adversarial Network (STGAN) framework to improve anomaly detection in urban traffic systems by incorporating external factors and applying it to real-time data from Gothenburg’s traffic network.

Furthermore, we construct a dynamic, real-world digital twin simulation of Gothenburg's urban traffic network by representing real-time traffic cameras as nodes and roads as edges \cite{Zhang2022}. This detailed modeling approach enables a more accurate representation of spatial dependencies and traffic flow variations across the city. By applying the STGAN framework to this comprehensive model, we are able to capture both short-term and long-term traffic patterns more effectively than existing methods.

\section{Related Work}\label{SOTASection}

Detecting traffic anomalies is a critical challenge in modern traffic management and urban planning, where understanding the spatial and temporal evolution of traffic is essential for effective decision-making \cite{Patil2022}. Machine learning approaches, particularly neural networks and generative models, have been widely applied to this problem \cite{Sakurada2015, Wang2019, norlander2019latent, astrom2024improved}. However, existing methods often fail to fully capture the complexity of urban traffic systems, which are influenced by both temporal dynamics and spatial topology \cite{Zhang2022}.

Long Short-Term Memory (LSTM) networks \cite{Hochreiter1997} have been extensively used for traffic forecasting due to their ability to model long-term temporal dependencies. Prior research on the Gothenburg dataset primarily employed LSTM networks to model temporal dynamics from individual cameras \cite{Sopasakis2019}. While effective in capturing temporal patterns, these methods overlook the crucial spatial dependencies between traffic nodes, which are essential for accurately predicting traffic flow \cite{Yu2018, Li2018}.

Graph Neural Networks (GNNs) provide a natural extension to traffic forecasting by explicitly modeling spatial dependencies between traffic nodes. GNNs have shown great potential in capturing spatial correlations and the influence of nearby traffic conditions \cite{Wu2020, Zhang2022}. For instance, Diffusion Convolutional Recurrent Neural Networks (DCRNN) effectively model spatiotemporal traffic data by integrating GNNs with recurrent neural networks \cite{Li2018, Yu2018}. However, many GNN-based methods inadequately incorporate temporal dynamics, which are essential for understanding traffic evolution and detecting anomalies arising due to spatiotemporal interactions \cite{Zhou2020}.

To address these limitations, recent research has focused on integrating GNNs with temporal models within a generative framework. The Spatiotemporal Generative Adversarial Network (STGAN) framework exemplifies this approach by jointly modeling spatial and temporal dependencies in traffic data \cite{Deng2022}. STGAN employs an adversarial architecture \cite{Wang2021} where a generator synthesizes plausible traffic sequences and a discriminator distinguishes them from real data, enhancing predictive accuracy and providing a robust mechanism for anomaly detection \cite{Goodfellow2014, Zhang2022}.

Our research builds upon these works by adopting the STGAN framework and applying it to real-time traffic data from Gothenburg. Integrating GNNs with LSTMs within a generative adversarial framework, allows to more effectively capture the complex spatiotemporal dynamics of urban traffic. This approach may improve anomaly detection accuracy by addressing some limitations of previous methods that either overlooked spatial dependencies or did not fully model temporal dynamics.

\section{Methods}\label{MethodsSection}

In this section, we present the framework of STGAN, building upon \cite{Deng2022}. We aim to enhance anomaly detection in traffic data by effectively capturing both spatial and temporal dependencies.

\subsection{Traffic Network Representation}

A traffic network is represented as a weighted graph $G = (V, E, \mathbf{W})$, where $V$ is a finite set of $N$ nodes (traffic cameras), $E$ is the set of edges representing connections between nodes, and $\mathbf{W} \in \mathbb{R}^{N \times N}$ is the weighted adjacency matrix representing spatial correlations.

The adjacency between nodes is determined based on their spatial proximity. Two nodes $v_i$ and $v_j$ are considered adjacent if the distance between them is within a predefined threshold or if they are connected by a road segment. The edge weight between nodes $v_i$ and $v_j$ is given by:

\[
\mathbf{W}_{ij} =
\begin{cases}
\exp \left(-\dfrac{\text{dist}(v_i,v_j)^2}{\sigma^2} \right), & \text{if } e_{ij} = 1, \\
0, & \text{otherwise},
\end{cases}
\]
where $\text{dist}(v_i,v_j)$ denotes the Euclidean distance between nodes $v_i$ and $v_j$, $\sigma$ is the standard deviation of all pairwise distances between nodes, and $e_{ij}$ indicates adjacency.

\subsection{Data Representation and Problem Definition}

Each data point is represented as the triplet $s = \langle v, t, \mathbf{x}_{v,t} \rangle$, where $v \in V$ is the node index, $t$ is the time index, and $\mathbf{x}_{v,t} \in \mathbb{R}^F$ is the feature vector at node $v$ and time $t$.

The historical data over $T$ time steps is denoted as $\mathcal{S} \in \mathbb{R}^{T \times N \times F}$, representing the traffic dynamics of all nodes. Given the traffic network $G$ and historical data $\mathcal{S}$, our goal is to identify anomalies at the next time step $T+1$ by predicting expected traffic measurements and detecting deviations indicative of anomalies.

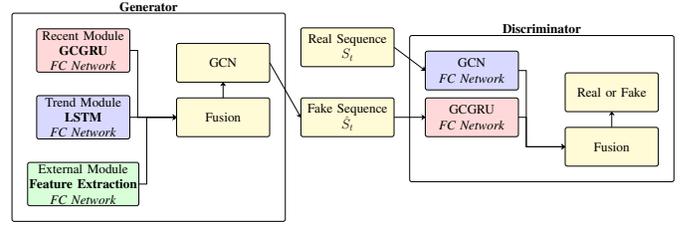
\begin{figure}
    \centering
    \resizebox{\linewidth}{!}{
\begin{tikzpicture}[
    font=\Huge,
    module/.style={
        rectangle, 
        draw, 
        rounded corners, 
        minimum width=6cm, 
        minimum height=2.5cm, 
        text centered, 
        align=center
    },
    module_red/.style={
        rectangle, 
        draw, 
        rounded corners, 
        fill=red!15, 
        minimum width=6cm, 
        minimum height=2.5cm, 
        text centered, 
        align=center
    },
    module_blue/.style={
        rectangle, 
        draw, 
        rounded corners, 
        fill=blue!15, 
        minimum width=6cm, 
        minimum height=2.5cm, 
        text centered, 
        align=center
    },
    module_green/.style={
        rectangle, 
        draw, 
        rounded corners, 
        fill=green!15, 
        minimum width=6cm, 
        minimum height=2.5cm, 
        text centered, 
        align=center
    },
    block/.style={
        rectangle, 
        draw, 
        rounded corners, 
        fill=yellow!20, 
        minimum width=6cm, 
        minimum height=2.5cm, 
        text centered, 
        align=center
    },
    arrow/.style={
        -{Stealth[length=6pt, width=8pt]}, 
        thick
    },
    generator_group/.style={
        draw, 
        rounded corners, 
        thick, 
        inner sep=1cm
    },
    discriminator_group/.style={
        draw, 
        rounded corners, 
        thick, 
        inner sep=1cm
    },
    node distance=1cm and 1cm,
]

\node (recent) [module_red, xshift=-8cm, yshift=3cm] {Recent Module \\ \textbf{GCGRU} \\ \textit{FC Network}};
\node (trend) [module_blue, below=of recent, yshift=-0.5cm] {Trend Module \\ \textbf{LSTM} \\ \textit{FC Network}};
\node (external) [module_green, below=of trend, yshift=-0.5cm] {External Module \\ \textbf{Feature Extraction} \\ \textit{FC Network}};

\node (fusion) [block, right= of trend, xshift=2cm] {Fusion};
\node (gcn) [block, above= of fusion] {GCN};

\draw [arrow] (recent.east) -- ++(right:0.5cm) |- (fusion.west);
\draw [arrow] (trend.east) -- ++(right:0.5cm) |- (fusion.west);
\draw [arrow] (external.east) -- ++(right:0.5cm) |- (fusion.west);
\draw [arrow] (fusion.north) -- (gcn.south);

\node (real_seq) [block, right=of gcn, yshift=1cm, xshift=1cm] {Real Sequence \\ $S_t$};
\node (fake_seq) [block, below=of real_seq, yshift=-1cm] {Fake Sequence \\ $\hat{S}_t$};

\node (disc_recent) [module_red, right=of fake_seq, xshift=1cm] {GCGRU \\ \textit{FC Network}};
\node (disc_gcn) [module_blue, above=of disc_recent, yshift=-0.5cm] {GCN \\ \textit{FC Network}};

\node (disc_fusion) [block, right= of disc_recent, xshift=2cm, yshift=-1.9cm] {Fusion};
\node (real_fake) [block, above= of disc_fusion] {Real or Fake};

\draw [arrow] (disc_recent.east) -- ++(right:0.5cm) |- (disc_fusion.west);
\draw [arrow] (disc_gcn.east) -- ++(right:0.5cm) |- (disc_fusion.west);
\draw [arrow] (disc_fusion.north) -- (real_fake.south);

\draw [arrow] (gcn.east) -- (fake_seq.west);

\draw [arrow] (fake_seq.east) -- (disc_recent.west);
\draw [arrow] (real_seq.east) -- (disc_gcn.west);

\node [generator_group, fit=(recent) (trend) (external) (fusion) (gcn), 
    label=above:{\textbf{\Huge Generator}}] {};

\node [discriminator_group, fit=(disc_recent) (disc_gcn) (disc_fusion) (real_fake), 
    label=above:{\textbf{\Huge Discriminator}}] {};

\end{tikzpicture}
}
    \caption{The STGAN framework includes a spatiotemporal generator (left) and a spatiotemporal discriminator (right). The generator processes three types of data: recent data (short-term temporal patterns) via a GCGRU module, trend data (long-term temporal patterns) via an LSTM, and external factors (contextual features such as weather or events) via a feature extraction network. These inputs are fused and passed through a GCN to model spatial dependencies, generating a fake sequence, $\hat{S}_t$. The discriminator, using GCGRU and GCN modules, evaluates the generated sequence against real sequences, $S_t$, to distinguish real from fake, enabling robust modeling of spatiotemporal dynamics. See also \cite{Deng2022}.}
    \label{fig:stgan}
\end{figure}

As shown in Fig.~\ref{fig:stgan}, the STGAN framework comprises two primary components:

\begin{itemize}
    \item \textbf{Spatiotemporal Generator ($G_{\theta}$)}: Generates predicted sequences of traffic data.
    \item \textbf{Spatiotemporal Discriminator ($D_{\phi}$)}: Distinguishes between real and generated sequences.
\end{itemize}

\subsubsection{Spatiotemporal Generator Components}

The spatiotemporal generator $G_{\theta}$ consists of three modules designed to capture different aspects of the traffic data:
\begin{enumerate}
    \item \textbf{Recent Module}: Captures short-term spatiotemporal dependencies using a Graph Convolutional Gated Recurrent Unit (GCGRU).
    \item \textbf{Trend Module}: Learns long-term temporal patterns using an LSTM network.
    \item \textbf{External Module}: Incorporates external factors (e.g., time of day, day of the week) using a fully connected layer.
\end{enumerate}
The outputs of these modules are fused using a Graph Convolutional Network (GCN) layer to produce the final prediction $\hat{\mathbf{X}}_{v,t}$ for each node $v$ at time $t$.

\subsubsection{Graph Convolutional Gated Recurrent Unit (GCGRU)}

The GCGRU extends the traditional GRU by integrating graph convolution operations to capture spatial dependencies among nodes. The update equations for the GCGRU at time $t$ are:
\begin{align}
& \mathbf{r}_t = \sigma\left( \mathbf{W}_r * \mathbf{X}_t + \mathbf{U}_r * \mathbf{H}_{t-1} + \mathbf{b}_r \right), \\
& \mathbf{z}_t = \sigma\left( \mathbf{W}_z * \mathbf{X}_t + \mathbf{U}_z * \mathbf{H}_{t-1} + \mathbf{b}_z \right), \\
& \tilde{\mathbf{H}}_t = \tanh\left( \mathbf{W}_h * \mathbf{X}_t + \mathbf{U}_h * (\mathbf{r}_t \odot \mathbf{H}_{t-1}) + \mathbf{b}_h \right), \\
& \mathbf{H}_t = \mathbf{z}_t \odot \mathbf{H}_{t-1} + (1 - \mathbf{z}_t) \odot \tilde{\mathbf{H}}_t,
\end{align}
where $\mathbf{X}_t \in \mathbb{R}^{N \times F}$ represents the input features at time $t$, $\mathbf{H}_{t} \in \mathbb{R}^{N \times D}$ is the hidden state, and $\sigma(\cdot)$ denotes the sigmoid activation function. The graph convolution operation is denoted by $*$.

\subsection{Loss Functions}

The generator aims to produce sequences that are both realistic and close to the true data. Its loss function combines adversarial loss and reconstruction loss:
\begin{equation}
\mathcal{L}_G(\boldsymbol{\theta}) = \sum_s \left[ -\log(D_{\phi}(\hat{\mathbf{S}}_{v,t})) + \lambda_G \| G_{\theta}(v,t) - \mathbf{X}_{v,t} \|_2^2 \right],
\label{eq:generator_loss}
\end{equation}
where $D_{\phi}(\hat{\mathbf{S}}_{v,t})$ denotes the discriminator's probability that the generated sequence is real, and $\lambda_G$ balances the adversarial and reconstruction losses.
The discriminator seeks to correctly classify real and generated sequences. Its loss function is:
\begin{equation}
\mathcal{L}_D(\boldsymbol{\phi}) = \sum_s \left[ -\log(D_{\phi}(\mathbf{S}_{v,t})) - \log(1 - D_{\phi}(\hat{\mathbf{S}}_{v,t})) \right].
\label{eq:discriminator_loss}
\end{equation}

\subsection{Anomaly Score Calculation}

Anomalies are detected by comparing the generated data to the real data and assessing the discriminator's confidence. The anomaly score for a data point $s = \langle v, t, \mathbf{x}_{v,t} \rangle$ is computed as:
\begin{align}
s_G(v,t) &= \| G_{\theta}(v,t) - \mathbf{X}_{v,t} \|_2^2, \\
s_D(v,t) &= D_{\phi}(\mathbf{S}_{v,t}) - D_{\phi}(\hat{\mathbf{S}}_{v,t}), \\
\text{score}(v,t) &= s_G(v,t) + \lambda s_D(v,t),
\end{align}
where $\lambda$ balances the contributions of $s_G$ and $s_D$.

\subsection{Model Training Procedure}
The STGAN is trained using an adversarial learning process, where the generator and discriminator are updated alternately \cite{Goodfellow2014}. The training procedure is summarized in Algorithm~\ref{alg:training_procedure}.
\begin{algorithm}[H]
\caption{STGAN Training Procedure}
\label{alg:training_procedure}
\begin{algorithmic}[1]
\REQUIRE Training data $\{\mathbf{S}_{v,t}\}$, initial parameters $\boldsymbol{\theta}$ and $\boldsymbol{\phi}$, learning rates $\eta_G$, $\eta_D$
\WHILE{not converged}
    \STATE \textbf{Generator Update}:
    \STATE Generate sequences $\hat{\mathbf{S}}_{v,t} = G_{\theta}(v,t)$
    \STATE Compute generator loss $\mathcal{L}_G(\boldsymbol{\theta})$ using Eq.~\eqref{eq:generator_loss}
    \STATE Update generator parameters: $\boldsymbol{\theta} \leftarrow \boldsymbol{\theta} - \eta_G \nabla_{\boldsymbol{\theta}} \mathcal{L}_G(\boldsymbol{\theta})$
    \STATE \textbf{Discriminator Update}:
    \STATE Evaluate discriminator on real data: $D_{\phi}(\mathbf{S}_{v,t})$
    \STATE Evaluate discriminator on generated data: $D_{\phi}(\hat{\mathbf{S}}_{v,t})$
    \STATE Compute discriminator loss $\mathcal{L}_D(\boldsymbol{\phi})$ using Eq.~\eqref{eq:discriminator_loss}
    \STATE Update discriminator parameters:\\ \quad $\boldsymbol{\phi} \leftarrow \boldsymbol{\phi} - \eta_D \nabla_{\boldsymbol{\phi}} \mathcal{L}_D(\boldsymbol{\phi})$
\ENDWHILE
\end{algorithmic}
\end{algorithm}

\subsection{Implementation Details}
\subsubsection{Architecture and Training Parameters}
The model architecture and training parameters are crucial for its performance. For the recent module, we use two layers of GCGRU with a hidden dimension of 64, allowing the model to capture complex spatiotemporal dependencies. The trend module consists of two LSTM layers with a hidden dimension of 64 to model long-term temporal patterns. The external module processes the 31-dimensional encoded time features through a fully connected layer with an output dimension of 64.

During training, we set the learning rate for both the generator and discriminator to $0.001$ and use a batch size of 64. The model is trained for 6 epochs, which was determined to be sufficient for convergence based on validation performance. The hyperparameters $\lambda_G$ in Eq.~\eqref{eq:generator_loss} and $\lambda$ in the anomaly score calculation were both set to $1$ in our experiments.

Optimization is performed using the Adam optimizer \cite{Kingma2014}, which adapts the learning rate during training for faster convergence. All experiments were conducted using PyTorch and the code is available at \url{https://github.com/fgias/traffic-anomaly-detection}.

\subsubsection{Dataset and Preprocessing}

The dataset is derived from cameras distributed across the city of Gothenburg in Sweden, provided by Trafikverket. The video data comprises images captured at one-minute intervals. Each camera is identified by a unique ID and accompanied by GPS coordinates indicating its location and orientation.

The images are processed to compute the \textit{flow} metric $\text{flow}_{v,t}$ for camera $v$ at time $t$, representing the coverage of the roads targeted by the camera. The flow metric is calculated using:

\begin{equation}
\text{flow}_{v,t} = \frac{\text{number of detected vehicles}}{\text{maximum vehicle capacity}},
\end{equation}
where the number of detected vehicles is obtained using a vehicle detection algorithm (e.g., YOLOv5 \cite{YOLO2020}), and the maximum vehicle capacity is a predefined constant based on road characteristics.

Before feeding the data into the algorithm, another processing is performed in order to smooth the data, patch the missing minutes, and truncate the times before 4:53 am and after 9:00 pm. Data from April 1st, 2020, until November 13th, 2020, are used for training. The results are verified on 10 days from November 14th, 2020, to November 23rd, 2020.

First, the data from all the cameras used, 42 in our case, are concatenated along the time direction. The missing minute holes are filled by propagating the last valid observation to next valid, using the forward fill method. The 1-minute data points are then reduced to 5-minute intervals by averaging, in order to smooth the variations. The node distances, the node subgraphs $G_v$, and the time feature are calculated. The time feature, for the external module, in our case is represented as,
\begin{equation}
    E = [ \boldsymbol{O}_{\text{weekday}}; \boldsymbol{O}_{\text{hour}} ],
\end{equation}
where $\boldsymbol{O}_{\text{weekday}}$ is a one-hot vector of length 7 representing the day of the week, and $\boldsymbol{O}_{\text{hour}}$ is a one-hot vector of length 24 representing the hour of the day. Weather features were not considered for our dataset, and therefore the feature is 31-dimensional. Finally, since the cameras are active only between 4:53 am and 9:00 pm, data before 4:55 and data after 9:00 pm are truncated. The time feature has to be truncated in the same way. The data are then ready to be passed on to STGAN for training.

\section{Results}\label{ResultsSection}
\begin{figure}
    \centering
    \includegraphics[width=1\linewidth]{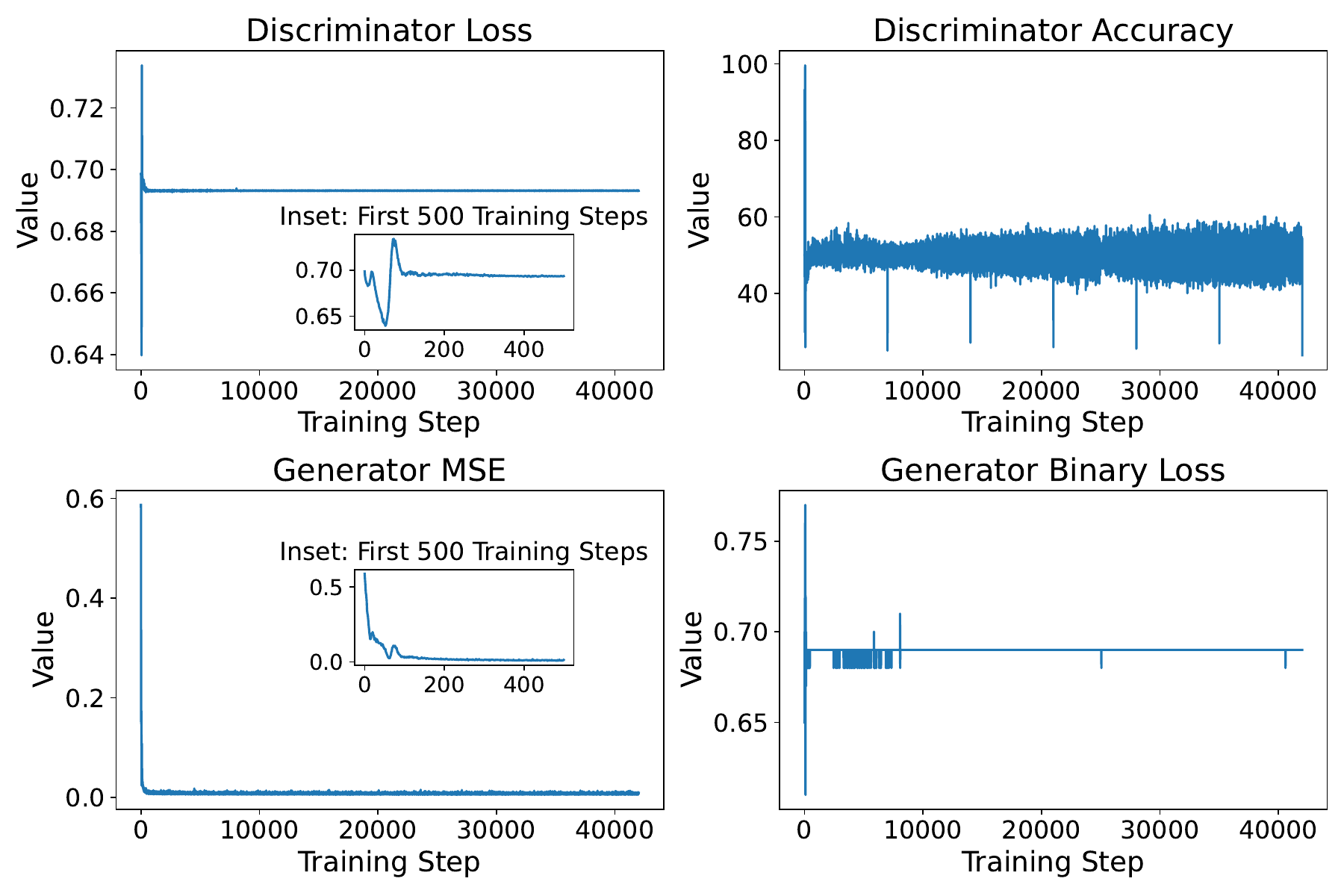}
    \caption{Training metrics for the adversarial training of the STGAN discriminator and generator.}
    \label{fig:training-plots}
\end{figure}

Training metrics such as discriminator and generator binary loss as well as discriminator accuracy and generator MSE, are presented in Figure~\ref{fig:training-plots}. The last 10 values of these metrics are summarized in Table~\ref{tab:training-metrics}. The low accuracies of the discriminator suggests that the generator has become good at generating fake sequences that are hard to distinguish from the real data.

Following \cite{Deng2022}, we calculate the anomaly scores of all the data in the test set. We label the data points with the top $K\%$ anomaly scores as anomalies. 

\begin{table}
\centering
\setlength{\tabcolsep}{4pt}
\caption{Discriminator and generator metrics for last 10 steps of training process.}
\begin{tabular}{cc|cccc}
\hline \hline
Epoch & Step & \makecell{Discriminator\\Loss} & \makecell{Discriminator\\Accuracy} & \makecell{Generator\\MSE} &\makecell{Generator\\Binary\\Loss} \\ \hline
6 & 6992 & 0.693070 & 53.12\% & 0.007648 & 0.693215 \\
6 & 6993 & 0.693185 & 43.95\% & 0.007306 & 0.693100 \\
6 & 6994 & 0.693192 & 48.44\% & 0.007066 & 0.693192 \\
6 & 6995 & 0.693146 & 48.44\% & 0.006912 & 0.693143 \\
6 & 6996 & 0.693195 & 45.12\% & 0.007161 & 0.693341 \\
6 & 6997 & 0.693172 & 49.80\% & 0.008960 & 0.693488 \\
6 & 6998 & 0.693201 & 51.17\% & 0.008466 & 0.693458 \\
6 & 6999 & 0.693100 & 54.49\% & 0.008442 & 0.693006 \\
6 & 7000 & 0.693140 & 47.27\% & 0.008975 & 0.692538 \\
6 & 7001 & 0.693116 & 23.83\% & 0.009361 & 0.692604 \\
\hline \hline
\end{tabular}
\label{tab:training-metrics}
\end{table}

Evaluating anomaly detection in real-world scenarios remains an open challenge, as obtaining a complete set of ground truth data is difficult. Since we were unable to do this, we take the output of the algorithm and manually verify each flagged anomaly. The precision is calculated by:

\begin{equation}
\text{precision} = \frac{\text{true positives}}{\text{true positives} + \text{false positives}}.
\end{equation}
Table~\ref{tab:precision} presents the precision of the STGAN algorithm on the test dataset for different values of $K$. For low values of $K$ there are very few positives, all of which are true positives---anomalies whose causes we were able to identify. As we increase $K$, we have a larger number of anomalies, some of which, however, are false positives, meaning that we were unable to explain their origin.

Table~\ref{tab:anomaly_types} shows the various types of anomalies detected for a specific value of $K$, classifying them into three anomaly types. The first type, ``camera signal cut/restart'', refers to signals due to problems with the functioning of the camera, such as the cutting or restarting of the signal, power outages etc. The second type, ``visual artifacts'', refers to anomalies that were triggered due to the visual quality of the input, such as the degrading of the clarity of the image due to moisture/droplets from rain on the camera lens, or the intense reflection of light on wet roads. Anomalies in this category were thus not identified with changes in the traffic flow. Lastly, ``extreme weather conditions'' is the type of anomalies which are due to developments in the weather that, in fact, produce disturbances in the traffic flow.

\begin{figure}
    \centering
    \includegraphics[width=.95\linewidth]{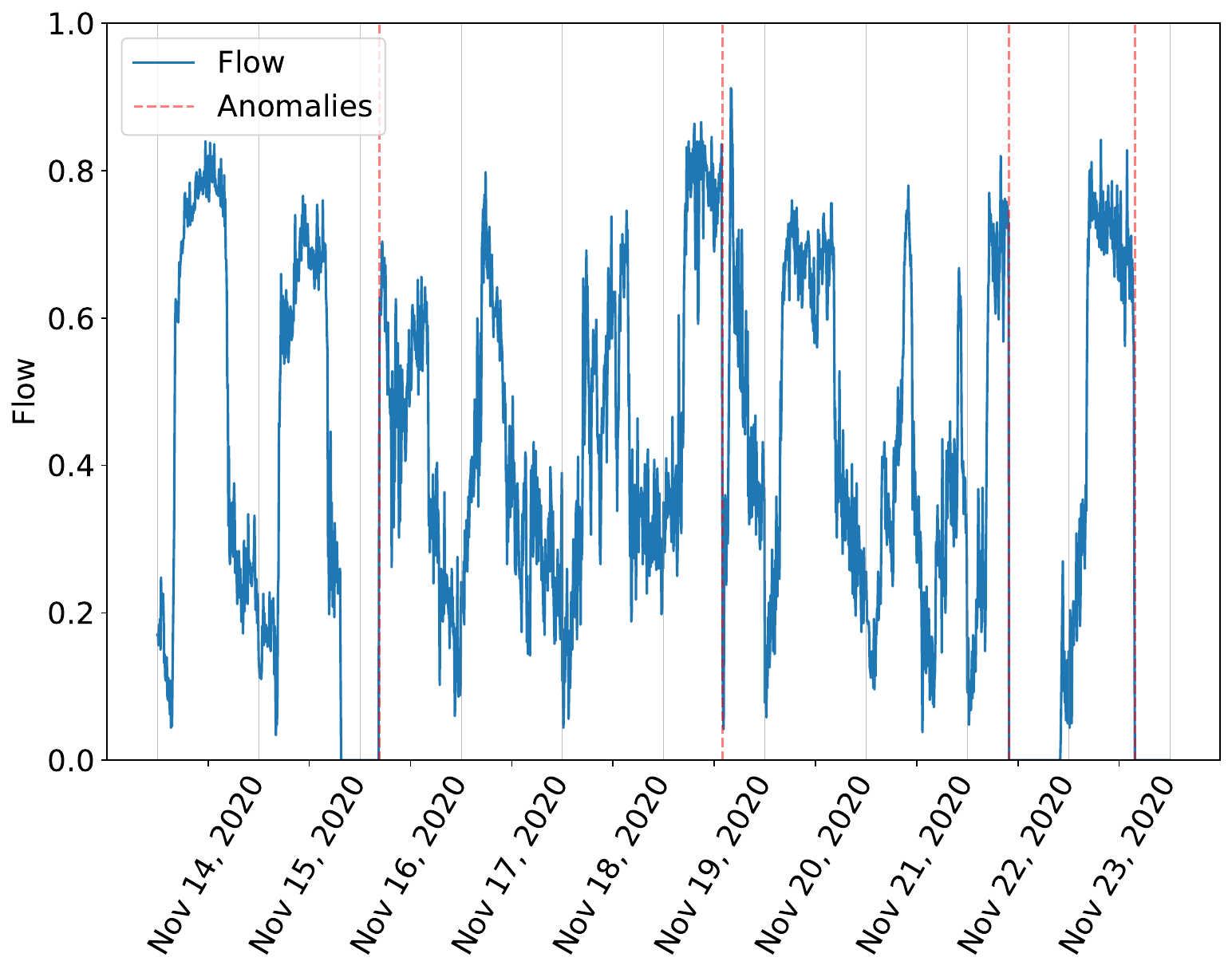}
    \caption{Anomalies detected for Camera 25, for 10 test days in November 2020, calculated for $K=0.1\%$.}
    \label{fig:anomalies}
\end{figure}

\begin{figure}
    \centering
    \includegraphics[width=.8\linewidth]{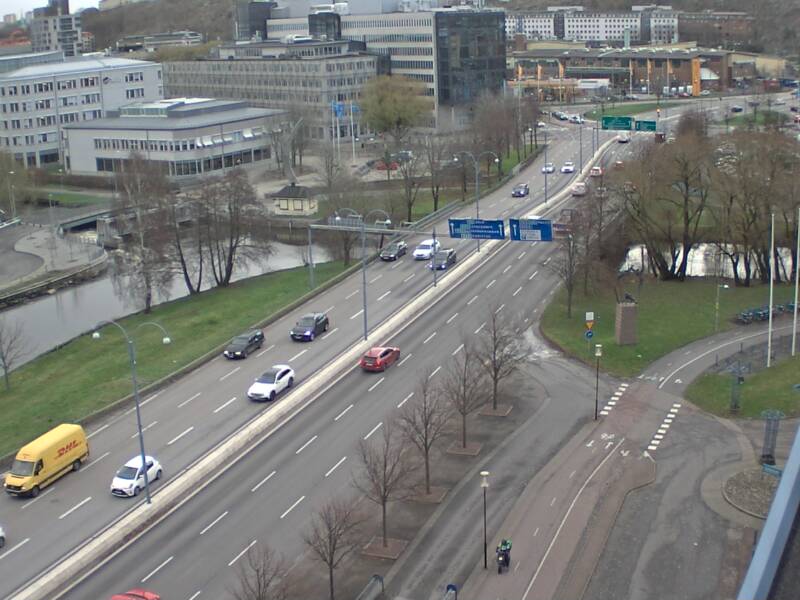}
    \vskip 0.5em 
    \includegraphics[width=.8\linewidth]{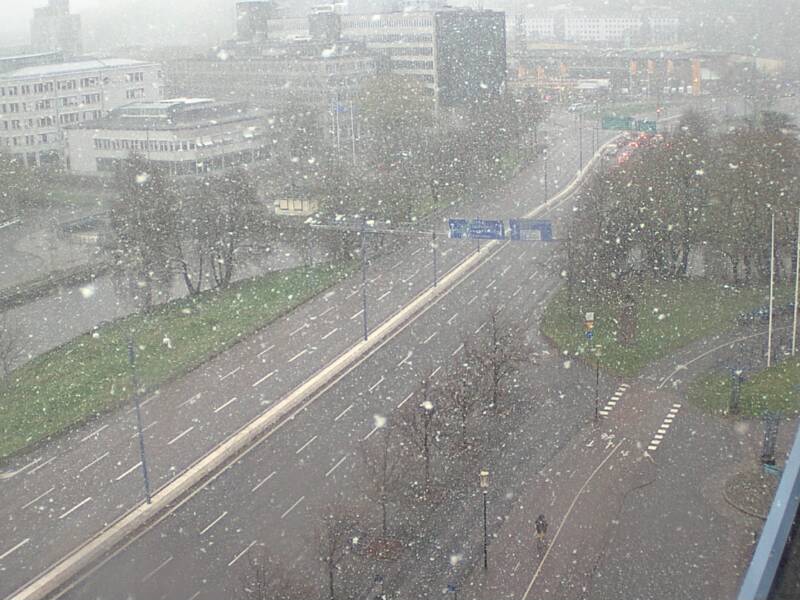}
    \caption{Detection of the beginning of heavy snowfall. Scenes from Camera 14 on Nov. 19, 2020, at 14:10 (top) and 14:20 (bottom).}
    \label{fig:heavy_snowfall}
\end{figure}

Fig.~\ref{fig:anomalies} illustrates the anomalies identified for Camera 25 across a 10-day test period in November 2020, with $K=0.1\%$. The anomalies, on November 16, 21 and 23, are anomalies of type ``camera signal cut/restart''. In the first case, the camera was not working during the normal function time window and triggered a signal when it started working. In the last two cases, the signal of the camera was cut earlier than expected, triggering anomalies. The most interesting case, however, is that of November 19. This anomaly is of type ``extreme weather conditions''. At the anomaly time, heavy snowfall started, which in turn disturbed the normal traffic flow, eventually triggering an anomaly. Fig.~\ref{fig:heavy_snowfall} shows the dramatic change of the weather between 14:10 and 14:20 from Camera 14.

\begin{table}
    \centering
    \caption{Precision of STGAN on the test dataset for various values of $K$.}
    \begin{tabular}{c|ccc|c}
    \hline \hline
    K (\%) & Positives & True Positives & False Positives & Precision (\%) \\ \hline
    0.01 & 8 & 8 & 0 & 100 \\
    0.025 & 20 & 20 & 0 & 100 \\
    0.05 & 40 & 40 & 0 & 100 \\
    0.075 & 61 & 59 & 2 & 96.7 \\
    0.1 & 81 & 75 & 6 & 92.6 \\
    \hline \hline
    \end{tabular}

    \label{tab:precision}
\end{table}

\begin{table}
    \centering
    \caption{Identified anomalies and anomaly types for $K=0.1\%$.}
    \begin{tabular}{c|c}
    \hline \hline
    Anomaly Type & Number of Anomalies \\ \hline
    Camera signal cut/restart & 71 \\
    Visual artifacts & 2 \\
    Extreme weather conditions & 2 \\
    \hline
    True positives & 75 \\
    False positives & 6 \\
    \hline
    Total & 81 \\
    \hline \hline
    \end{tabular}
    \label{tab:anomaly_types}
\end{table}

\section{Discussion}\label{DiscussionSection}

The results presented provide compelling evidence of the effectiveness of the STGAN framework for traffic anomaly detection. Table~\ref{tab:training-metrics} illustrates the performance metrics of both the generator and discriminator during training. The relatively stable loss values for both components indicate that the model is effectively learning to generate realistic traffic sequences while distinguishing them from actual data.

The precision values reported in Table~\ref{tab:precision} further substantiate our model’s efficacy. Achieving a precision of 100\% at lower thresholds demonstrates that our method can accurately identify anomalies without generating false positives. However, as $K$ increases, a slight drop in precision highlights a common trade-off in anomaly detection systems—balancing sensitivity and specificity.

The diversity of identified anomalies outlined in Table~\ref{tab:anomaly_types} reveals critical insights into the nature of traffic disruptions. The predominance of “camera signal cut/restart” anomalies suggests that infrastructure reliability is a significant concern in urban traffic monitoring. Additionally, the presence of visual artifacts and extreme weather conditions as contributing factors emphasizes the need for adaptive models capable of distinguishing between genuine traffic anomalies and environmental influences.

The results reveal that weather conditions, such as rain and snow, substantially impact the flow metrics derived from traffic camera observations. For instance, during adverse weather, visual artifacts can mislead traffic detection algorithms, resulting in erroneous vehicle density counts. Our model adeptly identifies these anomalies, showcasing its robustness against environmental variations.

Furthermore, recent studies, such as \cite{Zhang2022}, highlight the efficacy of spatial-temporal graph neural networks in traffic anomaly detection. Their work emphasizes the importance of modeling both spatial and temporal features; however, they do not fully leverage the generative capabilities of adversarial networks to enhance anomaly detection accuracy. Our approach not only models these features but also incorporates external factors like weather conditions and special events, which are often neglected in other studies.

\section{Conclusions}\label{ConclusionsSection}

Our findings indicate that the STGAN framework effectively captures complex spatiotemporal patterns in traffic data, leading to improved anomaly detection performance compared to traditional methods.

When compared to existing methodologies, utilizing LSTM networks alone \cite{Sopasakis2019}, the STGAN framework demonstrates superior accuracy in capturing both spatial dependencies and temporal dynamics. For example, while LSTMs can effectively model long-term temporal dependencies, they often overlook critical spatial relationships between traffic nodes \cite{Yu2018}. In contrast, the integration of GNNs allows for a more comprehensive representation of urban traffic networks.

While the STGAN framework demonstrates significant improvements in traffic anomaly detection, several limitations should also be noted. First, incorporating  external temporal features, such as time-of-day and day-of-week, does not account for other influential factors like special events, or roadworks, which can significantly impact traffic flow and anomaly detection performance. Additionally, the reliance on the flow metric derived from vehicle detection algorithms introduces dependency on the accuracy of these preprocessing steps; inaccuracies or biases in vehicle detection could propagate through the model, affecting its robustness. The absence of comprehensive ground truth labels for anomalies necessitated manual verification, which may introduce subjectivity and limits the scalability of the evaluation process. Finally, the computational complexity of the STGAN framework is also a factor to be considered for real-time deployment in large-scale traffic networks, highlighting the need for optimization to enhance efficiency without compromising accuracy.

Our findings not only validate the STGAN framework’s potential for real-time traffic anomaly detection but also point toward areas for future improvement, such as enhancing model robustness against environmental factors and optimizing hyperparameter selection for different urban contexts.

\section*{Acknowledgments}
This work is part of the SMARTHEP network, funded by the European Union’s Horizon 2020 research and innovation programme under Grant Agreement No. 956086. The work of A.~S. is also partially supported by grants from eSSENCE No. 138227, FORMAS No. 2022-151862, Rymdstyrelsen No. 2022-00282 as well as AgTech Sweden. The computations were enabled by resources provided by the National Academic Infrastructure for Supercomputing in Sweden (NAISS), partially funded by the Swedish Research Council through Grant Agreement No. 2022-06725. The authors would like to thank Trafikverket and the city of Gothenburg for access to their traffic camera data. The authors also extend their gratitude to the LIP6 laboratory for granting access to their GPU cluster, where the majority of the model trainings were conducted. 

\bibliographystyle{IEEEtran}
\bibliography{IEEEabrv,references}

\begin{thebibliography}{10}
\providecommand{\url}[1]{#1}
\csname url@samestyle\endcsname
\providecommand{\newblock}{\relax}
\providecommand{\bibinfo}[2]{#2}
\providecommand{\BIBentrySTDinterwordspacing}{\spaceskip=0pt\relax}
\providecommand{\BIBentryALTinterwordstretchfactor}{4}
\providecommand{\BIBentryALTinterwordspacing}{\spaceskip=\fontdimen2\font plus
\BIBentryALTinterwordstretchfactor\fontdimen3\font minus \fontdimen4\font\relax}
\providecommand{\BIBforeignlanguage}[2]{{%
\expandafter\ifx\csname l@#1\endcsname\relax
\typeout{** WARNING: IEEEtran.bst: No hyphenation pattern has been}%
\typeout{** loaded for the language `#1'. Using the pattern for}%
\typeout{** the default language instead.}%
\else
\language=\csname l@#1\endcsname
\fi
#2}}
\providecommand{\BIBdecl}{\relax}
\BIBdecl

\bibitem{Zhao2019}
L.~Zhao, Y.~Song, C.~Zhang, Y.~Liu, P.~Wang, T.~Lin, M.~Deng, and H.~Li, ``T-gcn: A temporal graph convolutional network for traffic prediction,'' \emph{IEEE Transactions on Intelligent Transportation Systems}, vol.~21, no.~9, pp. 3848--3858, 2020.

\bibitem{Yin2022}
X.~Yin, G.~Wu, J.~Wei, Y.~Shen, H.~Qi, and B.~Yin, ``Deep learning on traffic prediction: Methods, analysis, and future directions,'' \emph{IEEE Transactions on Intelligent Transportation Systems}, vol.~23, no.~6, pp. 4927--4943, 2022.

\bibitem{sopasakis2016information}
A.~Sopasakis and M.~A. Katsoulakis, ``Information metrics for improved traffic model fidelity through sensitivity analysis and data assimilation,'' \emph{Transportation Research Part B: Methodological}, vol.~86, pp. 1--18, 2016.

\bibitem{Benarmas2023}
R.~B. Benarmas and K.~B. Bey, ``A deep learning-based framework for road traffic prediction,'' \emph{Journal of Supercomputing}, vol.~79, pp. 12\,345--12\,360, 2023.

\bibitem{Patil2022}
\BIBentryALTinterwordspacing
P.~Patil, ``Applications of deep learning in traffic management: A review,'' \emph{International Journal of Business Intelligence and Big Data Analytics}, vol.~16, pp. 1--16, 2022. [Online]. Available: \url{https://research.tensorgate.org/index.php/IJBIBDA/article/download/26/24}
\BIBentrySTDinterwordspacing

\bibitem{alperovich2008stochastic}
T.~Alperovich and A.~Sopasakis, ``Stochastic description of traffic flow,'' \emph{Journal of Statistical Physics}, vol. 133, no.~6, pp. 1083--1105, 2008.

\bibitem{Sakurada2015}
\BIBentryALTinterwordspacing
K.~Sakurada and T.~Okatani, ``Change detection from a street image pair using cnn features and superpixel segmentation.'' in \emph{BMVC}, X.~Xie, M.~W. Jones, and G.~K.~L. Tam, Eds.\hskip 1em plus 0.5em minus 0.4em\relax BMVA Press, 2015, pp. 61.1--61.12. [Online]. Available: \url{http://dblp.uni-trier.de/db/conf/bmvc/bmvc2015.html#SakuradaO15}
\BIBentrySTDinterwordspacing

\bibitem{Zhang2022}
\BIBentryALTinterwordspacing
H.~Zhang, S.~Zhao, R.~Liu, W.~Wang, Y.~Hong, and R.~Hu, ``Automatic traffic anomaly detection on the road network with spatial-temporal graph neural network representation learning,'' \emph{Wireless Communications and Mobile Computing}, vol. 2022, no.~1, p. 4222827, 2022. [Online]. Available: \url{https://onlinelibrary.wiley.com/doi/abs/10.1155/2022/4222827}
\BIBentrySTDinterwordspacing

\bibitem{Lv2015}
Y.~Lv, Y.~Duan, W.~Kang, Z.~Li, and F.-Y. Wang, ``Traffic flow prediction with big data: A deep learning approach,'' \emph{IEEE Transactions on Intelligent Transportation Systems}, vol.~16, no.~2, pp. 865--873, 2015.

\bibitem{Deng2022}
L.~Deng \emph{et~al.}, ``Graph convolutional adversarial networks for spatiotemporal anomaly detection,'' \emph{IEEE Transactions on Neural Networks and Learning Systems}, vol.~33, no.~6, pp. 2416--2428, 2022.

\bibitem{Wu2020}
\BIBentryALTinterwordspacing
Z.~Wu, S.~Pan, F.~Chen, G.~Long, C.~Zhang, and P.~S. Yu, ``A comprehensive survey on graph neural networks,'' \emph{IEEE Transactions on Neural Networks and Learning Systems}, vol.~32, no.~1, pp. 4--24, 2021. [Online]. Available: \url{https://arxiv.org/abs/1901.00596}
\BIBentrySTDinterwordspacing

\bibitem{Hochreiter1997}
S.~Hochreiter and J.~Schmidhuber, ``Long short-term memory,'' \emph{Neural Computation}, vol.~9, no.~8, pp. 1735--1780, 1997.

\bibitem{Sopasakis2019}
\BIBentryALTinterwordspacing
A.~Sopasakis, ``Traffic demand and longer term forecasting from real-time observations,'' in \emph{Proceedings of the International Conference on Time Series and Forecasting (ITISE 2019), Volume 2}, J.~M. Corchado \emph{et~al.}, Eds.\hskip 1em plus 0.5em minus 0.4em\relax ITISE, 2019, pp. 345--356. [Online]. Available: \url{https://itise.ugr.es/ITISE2019_Vol2.pdf}
\BIBentrySTDinterwordspacing

\bibitem{Yu2018}
B.~Yu, H.~Yin, and Z.~Zhu, ``Spatio-temporal graph convolutional networks: A deep learning framework for traffic forecasting,'' in \emph{Proceedings of the 27th International Joint Conference on Artificial Intelligence (IJCAI)}, 2018, pp. 3634--3640.

\bibitem{Wang2019}
Y.~Wang, Y.~Zhang, X.~Piao, H.~Liu, and K.~Zhang, ``Traffic data reconstruction via adaptive spatial-temporal correlations,'' \emph{IEEE Transactions on Intelligent Transportation Systems}, vol.~20, pp. 1531--1543, 2019.

\bibitem{norlander2019latent}
\BIBentryALTinterwordspacing
E.~Norlander and A.~Sopasakis, ``Latent space conditioning for improved classification and anomaly detection,'' \emph{arXiv preprint arXiv:1911.10599}, 2019. [Online]. Available: \url{https://arxiv.org/abs/1911.10599}
\BIBentrySTDinterwordspacing

\bibitem{astrom2024improved}
\BIBentryALTinterwordspacing
O.~{\AA}str{\"o}m and A.~Sopasakis, ``Improved anomaly detection through conditional latent space vae ensembles,'' \emph{arXiv preprint arXiv:2410.12328}, 2024. [Online]. Available: \url{https://arxiv.org/abs/2410.12328}
\BIBentrySTDinterwordspacing

\bibitem{Li2018}
Y.~Li, R.~Yu, C.~Shahabi, and Y.~Liu, ``Diffusion convolutional recurrent neural network: Data-driven traffic forecasting,'' in \emph{International Conference on Learning Representations (ICLR)}, 2018.

\bibitem{Zhou2020}
J.~Zhou, G.~Cui, S.~Hu, Z.~Zhang, C.~Yang, Z.~Liu, L.~Wang, C.~Li, and M.~Sun, ``Graph neural networks: A review of methods and applications,'' \emph{AI Open}, vol.~1, pp. 57--81, 2020.

\bibitem{Wang2021}
\BIBentryALTinterwordspacing
P.~Wang, Y.~Zhang, S.~Wang, L.~Li, and X.~Li, ``Forecasting travel speed in the rainfall days to develop suitable variable speed limits control strategy for less driving risk,'' \emph{Journal of Advanced Transportation}, vol. 2021, no.~1, p. 6639559, 2021. [Online]. Available: \url{https://onlinelibrary.wiley.com/doi/abs/10.1155/2021/6639559}
\BIBentrySTDinterwordspacing

\bibitem{Goodfellow2014}
\BIBentryALTinterwordspacing
I.~J. Goodfellow, J.~Pouget-Abadie, M.~Mirza, B.~Xu, D.~Warde-Farley, S.~Ozair, A.~Courville, and Y.~Bengio, ``Generative adversarial nets,'' in \emph{Advances in Neural Information Processing Systems 27 (NIPS)}, 2014, pp. 2672--2680. [Online]. Available: \url{https://papers.nips.cc/paper/5423-generative-adversarial-nets.pdf}
\BIBentrySTDinterwordspacing

\bibitem{Kingma2014}
\BIBentryALTinterwordspacing
D.~P. Kingma and J.~Ba, ``Adam: A method for stochastic optimization,'' \emph{arXiv preprint arXiv:1412.6980}, 2014. [Online]. Available: \url{https://arxiv.org/abs/1412.6980}
\BIBentrySTDinterwordspacing

\bibitem{YOLO2020}
\BIBentryALTinterwordspacing
G.~Jocher, ``{YOLOv5} by {Ultralytics},'' May 2020. [Online]. Available: \url{https://github.com/ultralytics/yolov5}
\BIBentrySTDinterwordspacing

\end{thebibliography}

\end{document}